\documentclass{article}
\usepackage{spconf,amsmath,graphicx}
\usepackage{epstopdf}
\usepackage{url}
\graphicspath{{./ICIP2017Picture/CNNArchitexture/}{./ICIP2017Picture/artifact/}{./ICIP2017Picture/lowcontrast/}{./ICIP2017Picture/Variaty/}}

\title{GLOBAL AND LOCAL INFORMATION BASED DEEP NETWORK FOR SKIN LESION SEGMENTATION}
\name{Jin Qi$^{\star\bigtriangledown\dagger}$\thanks{Thanks to the Center for Information in Medicine in the University of Electronic Science and Technology of China for Funding.},
Miao Le$^{\star}$, Chunming Li$^{\star}$ and Ping Zhou$^{\diamond\dagger}$\thanks{$^\dagger$Corresponding Author: Jin Qi (jqi@uestc.edu.cn); Ping Zhou (ping\_zhou950809@126.com)}}
\address{$^{\star}$Electrical Engineering Department,
          $^{\bigtriangledown}$Center for Information in Medicine, $^{\diamond}$Cancer Hospital/Center \\
in University of Electronic Science and Technology of China \\
        Xiyuan Avenue 2006, West Gaoxin District, Chengdu, Sichuan Province, 611731, China }
\begin{document}
%
\maketitle
\begin{abstract}
With a large influx of dermoscopy images and a growing shortage of dermatologists, automatic dermoscopic image analysis plays an essential role in skin cancer diagnosis. In this paper, a new deep fully convolutional neural network (FCNN) is proposed to automatically segment melanoma out of skin images by end-to-end learning with only pixels and labels as inputs. Our proposed FCNN is capable of using both local and global information to segment melanoma by adopting skipping layers. The public benchmark database consisting of 150 validation images, 600 test images and 2000 training images in the melanoma detection challenge 2017 at International Symposium Biomedical Imaging 2017 is used to test the performance of our algorithm. All large size images (for example, $4000\times 6000$ pixels)  are reduced to much smaller images with $384\times 384$ pixels (more than 10 times smaller).  We got and submitted preliminary results to the challenge without any pre or post processing.  The performance of our proposed method could be further improved by data augmentation and by avoiding image size reduction.

\end{abstract}
\begin{keywords}
Skin Lesion, Dermoscopy Image, Deep Learning, Fully Convolutional Neural Network, Segmentation.
\end{keywords}
\section{Introduction}
\label{sec:intro}

Over five million cases of skin cancer are newly diagnosed in the United States annually\cite{Siegel16-Nature}. Early detection of melanoma, one of the most lethal forms of skin cancer, is critical in finding curable melanomas as well as increasing survival rate\cite{Freedberg1999738,Charles2001}. With a large influx of dermoscopy images, arriving of inexpensive consumer dermatoscope attachments for smart phone\cite{MoleScope} and a growing shortage of dermatologists\cite{Kimball2008741}, automatic dermoscopic image analysis plays an essential role in timely skin cancer diagnosis. Especially, lesion segmentation is one of the most important steps in dermoscopy image access.   

\begin{figure}[htb]

\begin{minipage}[b]{.48\linewidth}
  \centering
  \centerline{\includegraphics[width=3.0cm,height=2cm]{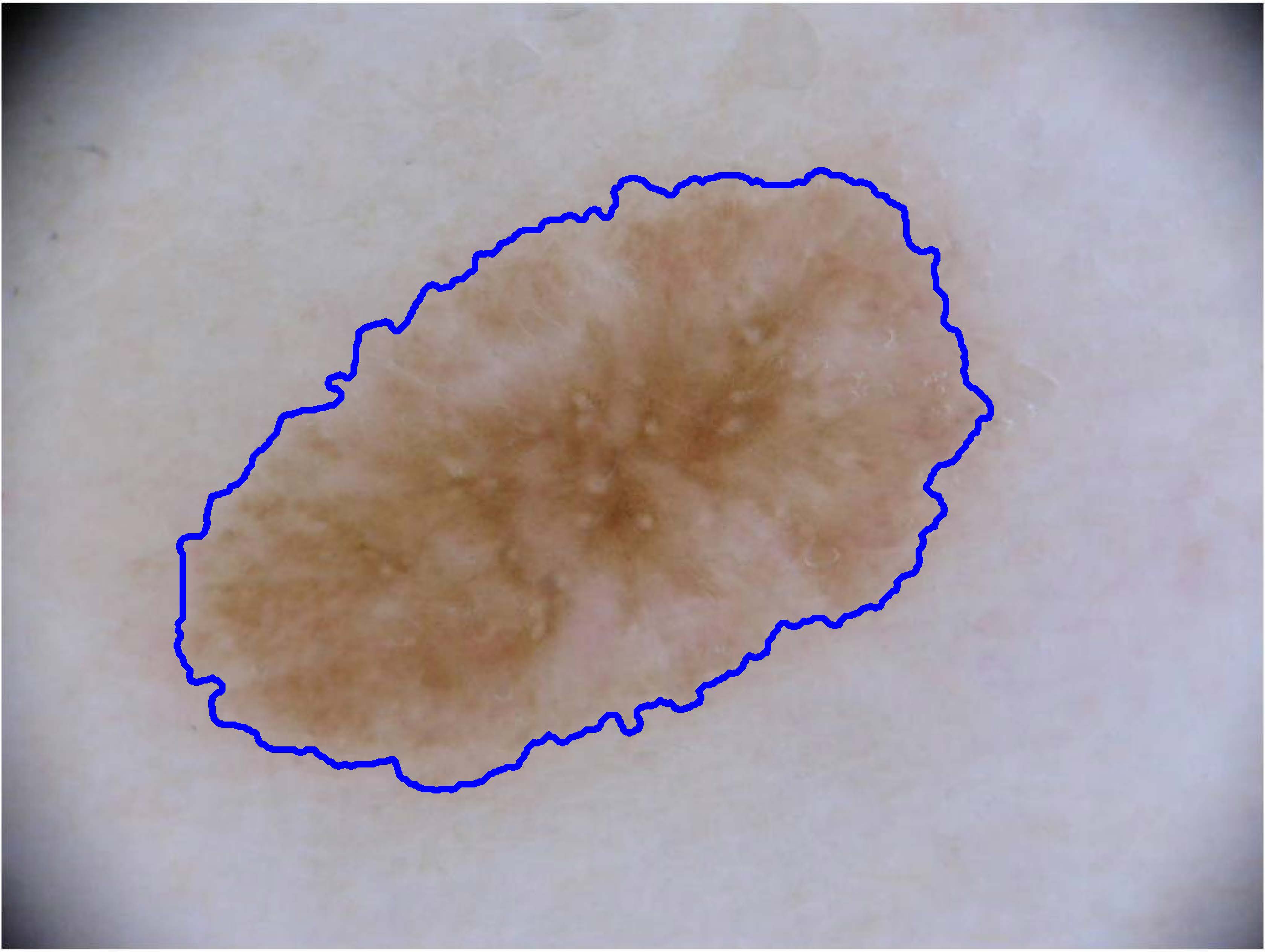}}
\end{minipage}
\begin{minipage}[b]{.48\linewidth}
  \centering
  \centerline{\includegraphics[width=3.0cm,height=2cm]{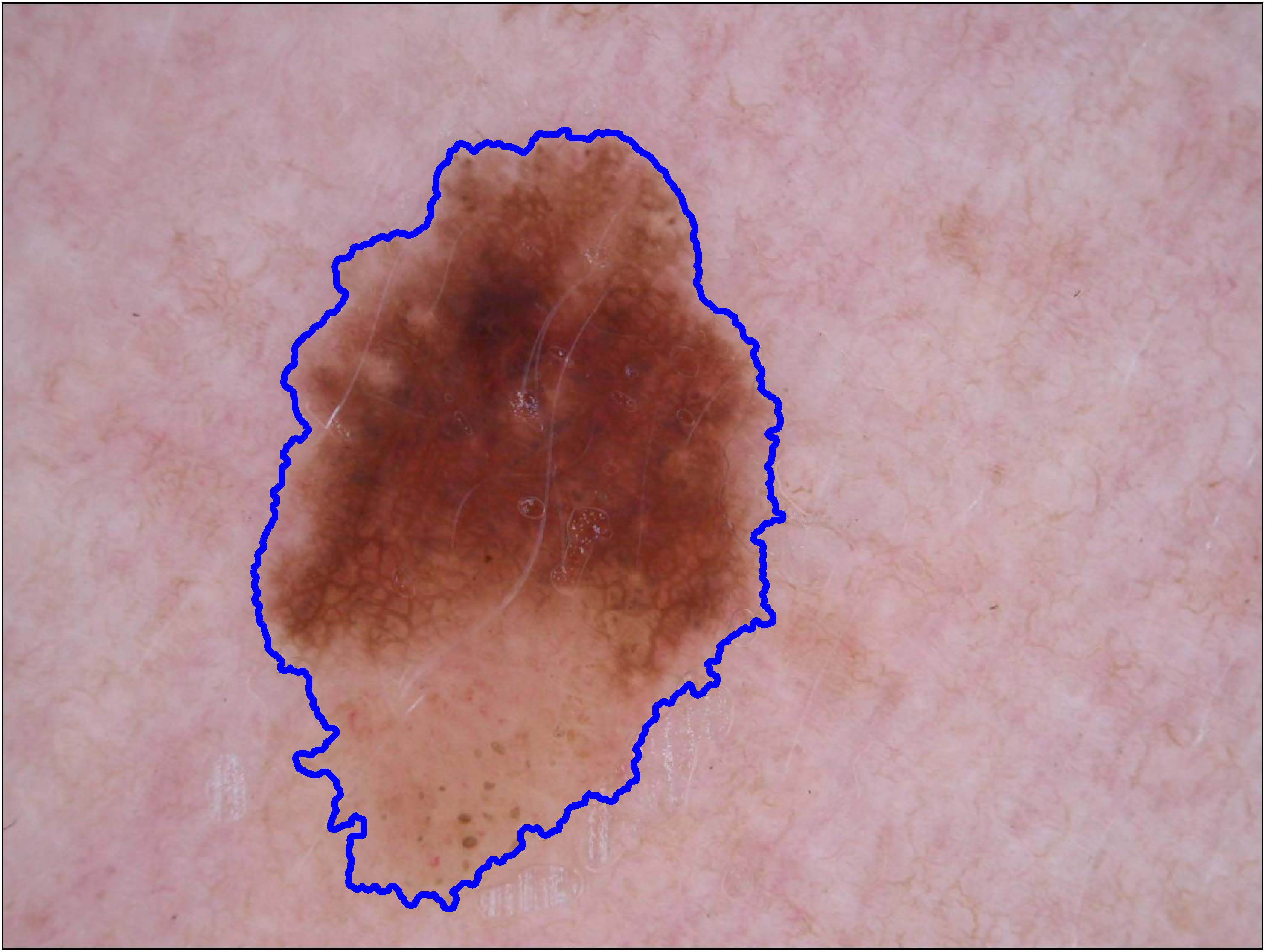}}
\end{minipage}
\hfill
\begin{minipage}[b]{0.48\linewidth}
  \centering
  \centerline{\includegraphics[width=3.0cm,height=2cm]{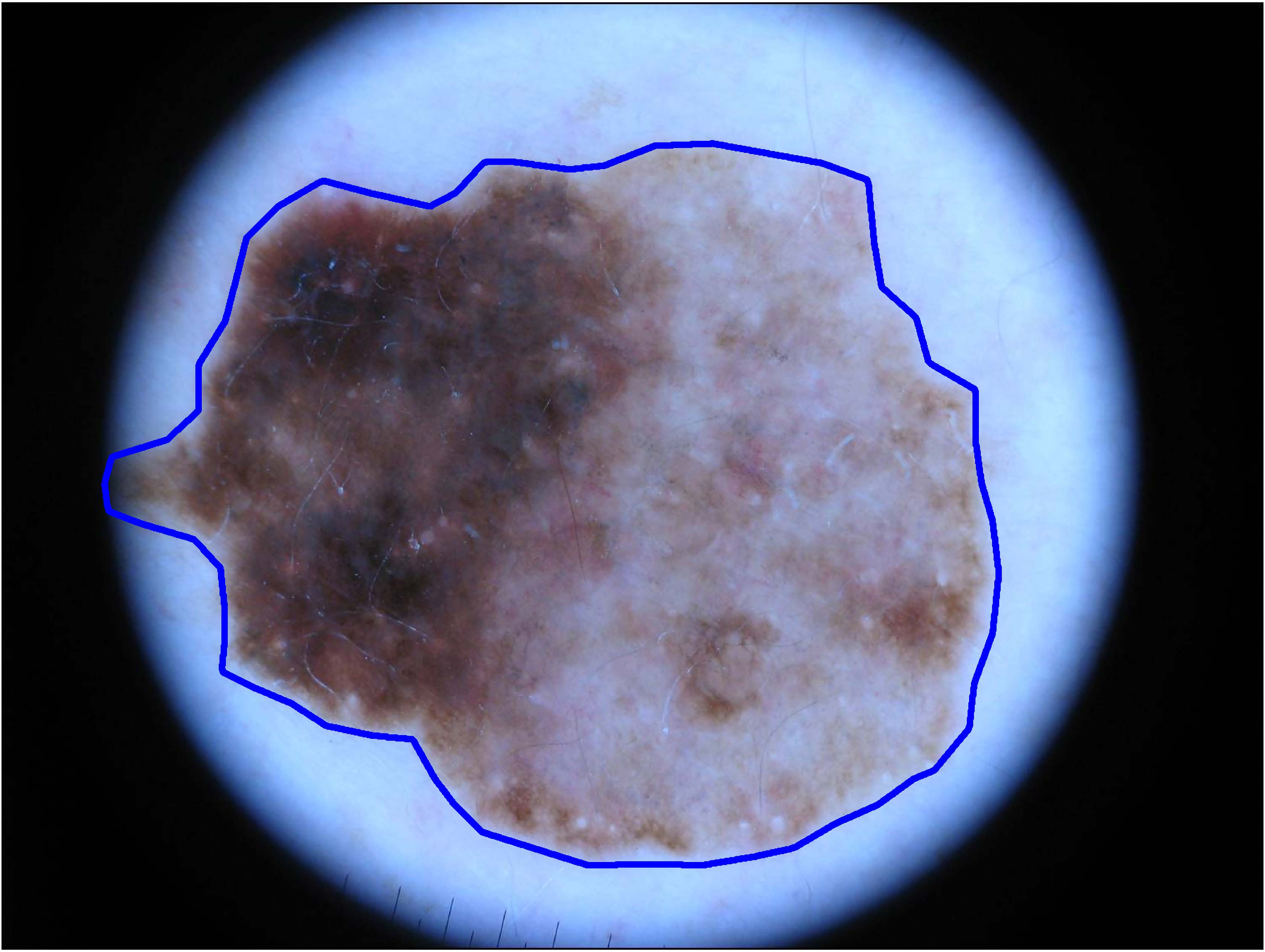}}
\end{minipage}
\begin{minipage}[b]{0.48\linewidth}
  \centering
  \centerline{\includegraphics[width=3.0cm,height=2cm]{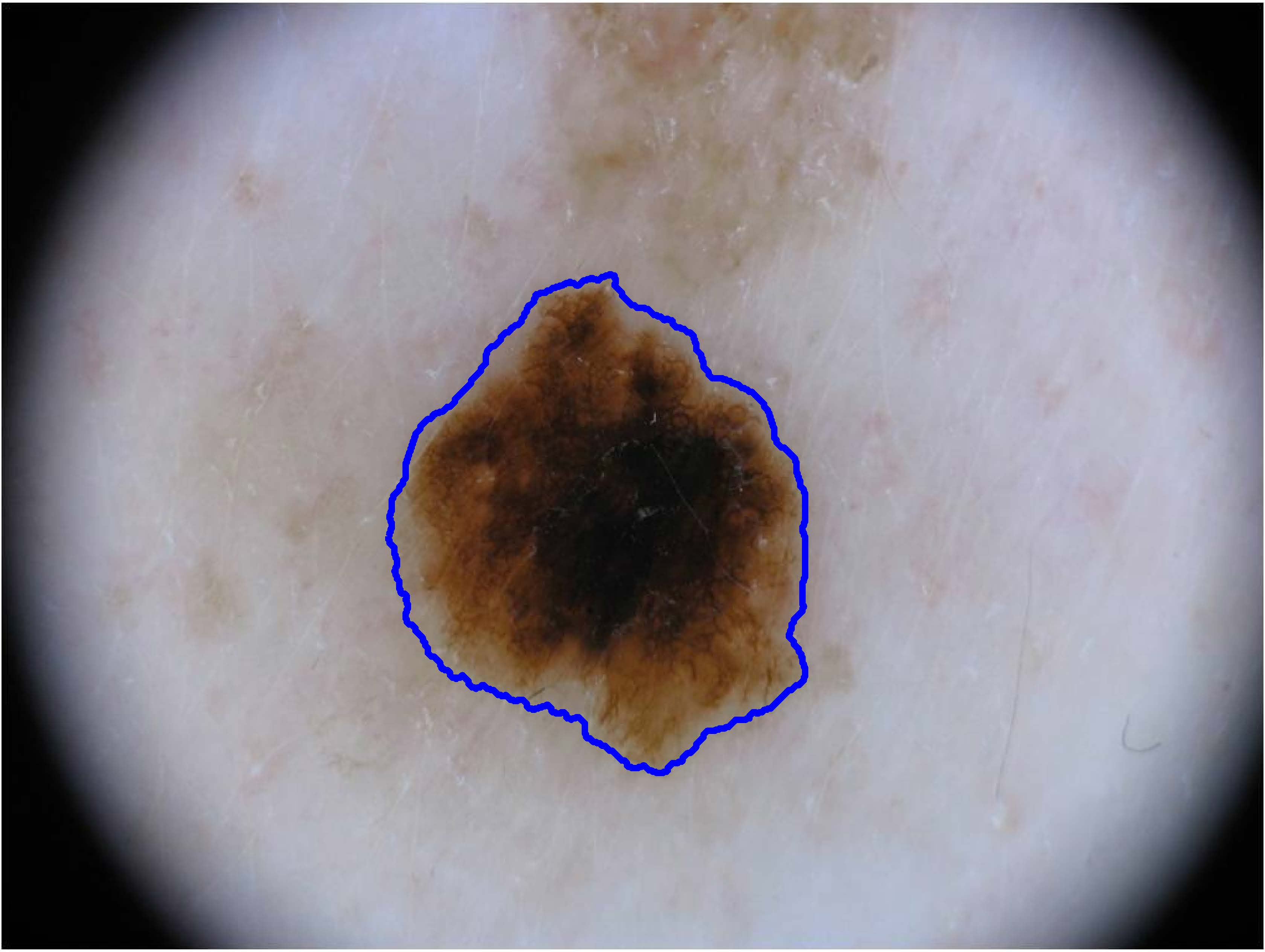}}
\end{minipage}
\begin{minipage}[b]{0.48\linewidth}
  \centering
  \centerline{\includegraphics[width=3.0cm,height=2cm]{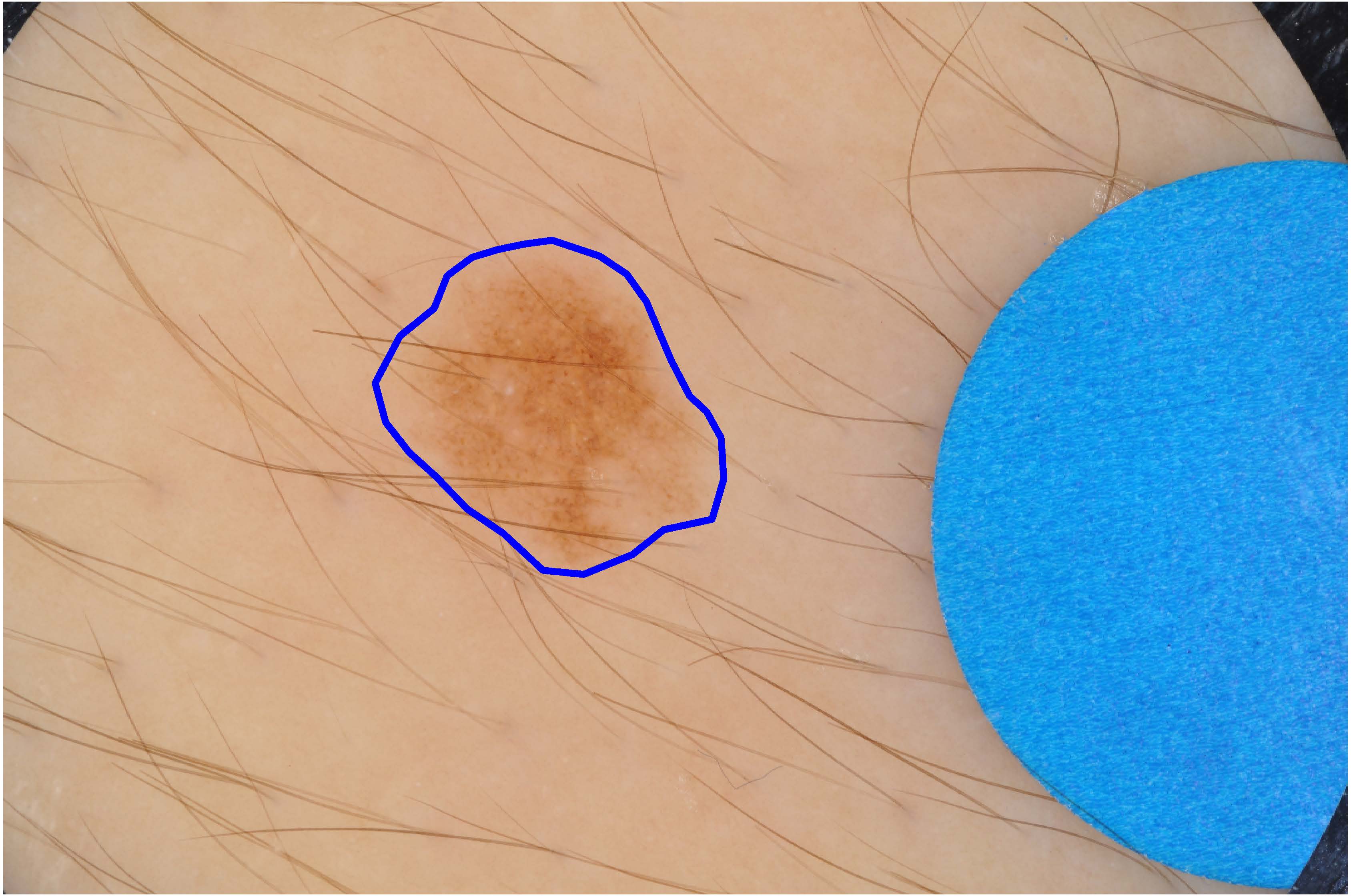}}
\end{minipage}
\hfill
\begin{minipage}[b]{0.48\linewidth}
  \centering
  \centerline{\includegraphics[width=3.0cm,height=2cm]{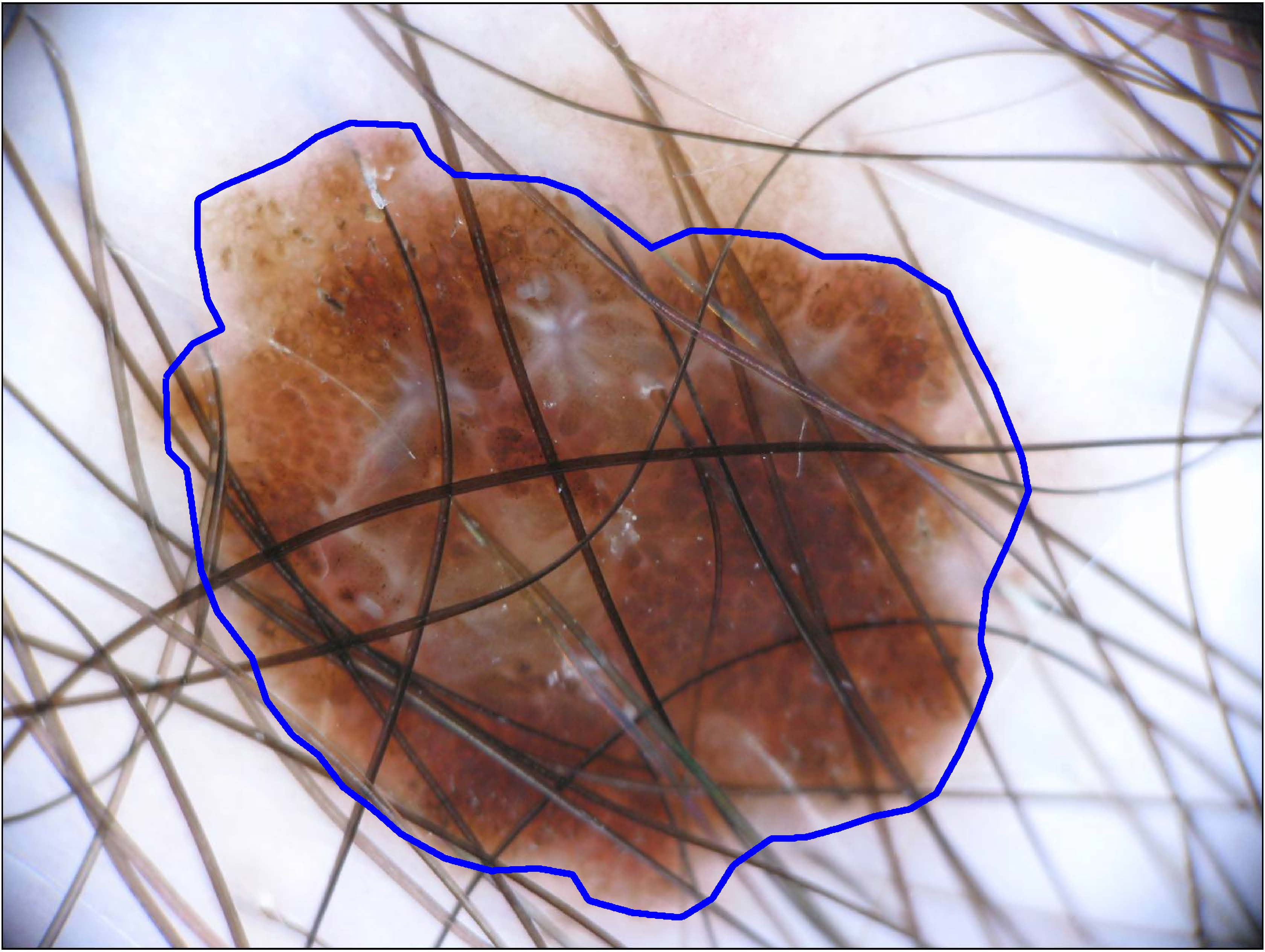}}
\end{minipage} 
\caption{Difficult example images with low contrast, large intraclass variety, and artifacts. Ground truth contours are marked in blue color. First row: low contrast between skin lesion and normal skin regions; Second row: large intraclass variance; Last row: artifacts in images. All images are from ISBI2017 SLATMDC\cite{2017-ISBI-SkinLesionChallenge}.}
\label{fig:example-artifacts}
\end{figure}

However, automatic lesion segmentation from dermoscopy images is a challenging task. The relatively low contrast and obscure boundaries between early stage skin lesions and normal skin regions make it very difficult to distinguish lesion area from normal skin region. The situation deteriorates rapidly when skin lesions are blurred or excluded by artifacts such as natural hairs, veins, artificial ruler marks, color calibration charts, and air bubbles, etc. Some example images from  
"ISBI 2017 Skin Lesion Analysis Towards Melanoma Detection Challenge" (ISBI2017 SLATMDC)\cite{2017-ISBI-SkinLesionChallenge} are shown in Fig.\ref{fig:example-artifacts}.

Many works have been done to try to deal with this difficult problem. Previously, some researchers proposed to segment lesions based on hand-crafted features\cite{Tommasi2006,Celebi2007362,Ganster2001,Schaefer2014,Jafari2016ICIP}. Recently, convolutional neural network(CNN) is adopted to improve medical image segmentation accuracy\cite{ronneberger2015unet,Chen2017135DCAN}. Deep CNN based skin cancer recognition has been initially proven to achieve dermatologist-level classification accuracy\cite{Esteva2017DermatologicstLevel-Nature}.  Fully convolutional neural network(FCNN) has shown promising segmentation results in natural images(PASCAL VOC) by fusing multi-scale prediction maps from multiple layers in itself\cite{Long2015FullyCN}. FCNN is based on a simple end-to-end learning and multi-scale contextual information is automatically explored in FCNN framework.

Inspired by \cite{Long2015FullyCN}, we propose a new FCNN framework for skin lesion segmentation in this paper. We pick up the VGG 16-layer net(publicly available from the Caffe model zoo) pretrained in ImageNet, discard its last classification layer, and replace all fully connected layers by convolution layers with randomly initialized weights. The weights in the remaining layers are kept to tackle small medical image dataset problem based on transfer learning. We add skipping layers to fuse multi-scale prediction maps from multiple layers. Different from \cite{Long2015FullyCN} which fuses prediction maps by pixel-wise plus operation, we treat multi-scale prediction maps as multi-scale feature maps and fuse them by concatenation operation. With our concatenating fusing strategy to keep fused features in higher space, we can capture more distinguishing local features and global features. Then we add a randomly initialized $1\times1$ convolution layer as the last layer to finish prediction with fused features as inputs. 

In \cite{Yu2016ResidualNet}, very deep fully convolutional residual network (FCRN) was proposed to do skin lesion segmentation. The authors in \cite{Yu2016ResidualNet} focused on residual network and very deep net(more than 50 layers). However, our proposed FCNN has different structures and is relatively shallower(less than 20 layers). Our experimental results are still comparable with that from \cite{Yu2016ResidualNet}. Our major contributions in this paper are summarized as follows:

\begin{itemize}
\item {We design a novel fully convolutional neural network for skin lesion segmentation in dermatoscopic images with end-to-end learning, transfer learning, and pixel by pixel prediction. }
\item{We treat multi-scale prediction maps as feature maps and propose a new fusing layer with a concatenation operation to combine these multi-scale contextual information to obtain better distinguishing features. We also add a convolution layer as the last layer of our FCNN to take the fused features as inputs to finish prediction.}
\item{We evaluate our proposed FCNN on the public dermatoscopy image dataset from  
"ISBI 2017 Skin Lesion Analysis Towards Melanoma Detection Challenge"\cite{2017-ISBI-SkinLesionChallenge}.
 We got and submitted preliminary results to the challenge without any pre or post processing.} 
\end{itemize}       

The remainder of this paper is organized as follows. We detail our method in Section \ref{sec:method}. Experimental results are reported in Section \ref{sec:experiment}. Section \ref{sec:discussion} is our discussion and conclusion.

\section{PROPOSED FULLY CONVOLUTIONAL NETWORK}
\label{sec:method}

\begin{figure*}[htb]

\begin{minipage}[b]{1.0\linewidth}
  \centering
  \centerline{\includegraphics[width=0.9\linewidth,height=5.5cm]{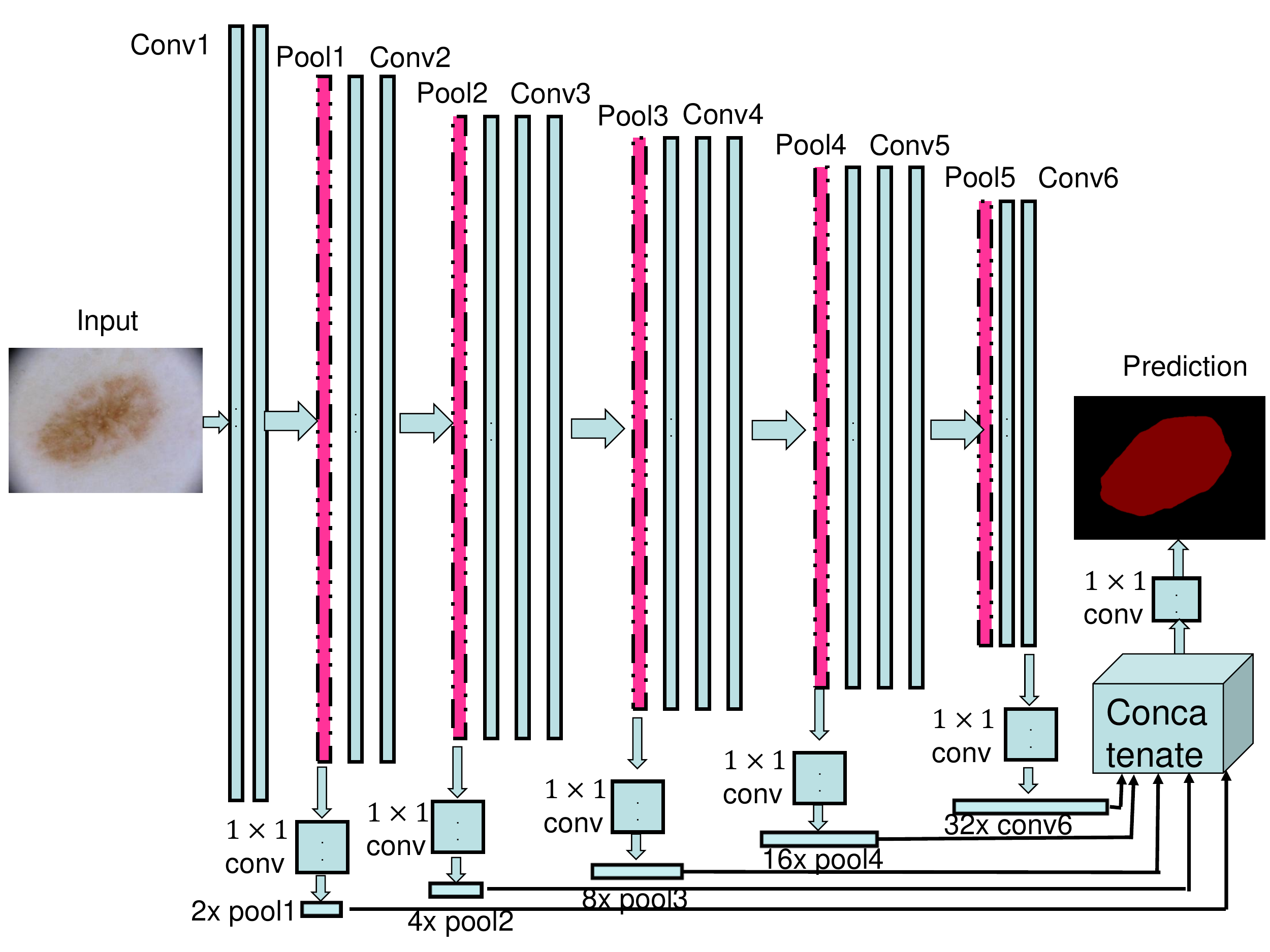}}
\end{minipage}
%
%
\caption{Architecture of our proposed FCNN. Blue vertical bar with solid contour: convolution layer; red vertical bar with dash contour: pooling layer; Blue small square: $1\times 1$ convolution layer; Blue horizontal bar: deconvolution layer for upsampling; Blue rectangular cube: fusing layer with concatenation operation.  Our FCNN learns to use multi-scale information.}
\label{fig:architecture-network}
\end{figure*}

In this section, we detail our proposed fully convolutional network for skin lesion segmentation. Due to relatively small number of medical images and labels available, we begin with pretrained network in image classification task to design our own dense prediction network in order to use transfer learning. Skip layers are added to incorporate multi-scale information into our network.    
The architecture of our proposed FCNN is illustrated in Fig. \ref{fig:architecture-network}.
\subsection{Basic Network Architecture}
 We pick up the pretrained VGG 16-layer net (publicly available from the Caffe model zoo) consisting of 13 convolution layers, 3 fully connected layers, 15 RELU layers, 5 pooling layers, 2 dropout layers and 1 softmax layer.  This VGG net has been trained for natural image classification in ImageNet. For our dense segmentation purpose, we discard its last two classification layers (last fully connected layer and softmax layer) and replace the remaining 2 fully connected layers with randomly initialized convolution layers. The learned weights in the remaining layers are kept to tackle small medical image dataset problem based on transfer learning. Thus, The modified network is a fully convolutional network which is able to take arbitrary size of images as inputs.

\subsection{Adding Skip Layers}
We add skipping layers to fuse multi-scale prediction maps from multiple layers. Different from \cite{Long2015FullyCN} which fuses prediction maps by pixel-wise sum operation, we treat multi-scale prediction maps as multi-scale feature maps and fuse them by concatenation operation. Specifically, after each pooling layer we add an $1\times1$ convolution layer to compute a dense prediction map with 2 channels (number of classes for each pixel in this paper). The $1\times1$ convolution layer can be considered as a classifier to classify each pixel independently\cite{Long2015FullyCN}.  An $1\times1$ convolution layer is also attached to the last convolution layer in basic network from above subsection. Thus we already added 6 $1\times1$ convolution layers in total to the basic network.  
The outputs of the 6 added $1\times1$ convolution layers are 6 dense prediction maps (2 channels each map).

The above 6 dense prediction maps have different resolutions or sizes. We add multiple deconvolution layers\cite{Long2015FullyCN} to upsample the 6 dense maps to the original size of the input image. The weights of deconvolution layers are randomly initialized and all of them can be learned automatically\cite{Long2015FullyCN}. Then we add a concatenation layer to concatenate the 6 upsampled prediction maps (with the same size as input image) in third dimension to form a final prediction feature map with $6\times2=12$ channels. 

With our concatenating fusing strategy to keep fused features in higher space (12 dimension in this paper), we can capture more distinguishing local features and global features. Then we add a randomly initialized $1\times1$ convolution layer after the above added concatenation layer as the last layer to compute the final prediction map with the concatenated feature map as input. In training stage, we add a softmax loss layer to fine-tune our network. In test, the loss layer is removed. 

\section{EXPERIMENTAL RESULTS}
\label{sec:experiment}

\subsection{Dataset}
We use the public dermatoscopy image dataset from  
"ISBI 2017 Skin Lesion Analysis Towards Melanoma Detection Challenge"\cite{2017-ISBI-SkinLesionChallenge} to evaluate the performance of our proposed method in this paper.
This dataset consists of 150 validation images, 600 test images and 2000 training images.
The ground truth binary mask images from expert manual segmentation are also provided  with pixel value 255 and 0 indicating lesion pixel and skin pixel, respectively. 


\subsection{Implementation}
Our proposed method is implemented with Maltlab by using the open source deep learning toolbox  "MatConvNet"\cite{vedaldi15matconvnet}. Our computer has a NVIDIA GTX1080 GPU card, Windows 7 system, an Intel i7-7700K processor with 4.5 GHz and 64G memory.  
In order to tackle the small medical data problem, we utilize a pretrained VGG 16-layer net (publicly available from the Caffe model zoo) to initialize some weights of our proposed FCNN based on transfer learning idea.  
Stochastic gradient descent(SGD) is used to fine-tune our proposed FCNN with batch size 6, momentum 0.9, weight decay 0.0001, learning rate 0.001.  

\subsection{Evaluation Metrics}
We also use the same metrics as the challenge organizer to evaluate the performance of our proposed method. These metrics are sensitivity(SE), specificity(SP), accuracy(AC), Jaccard index(JA) and Dice coefficient(DI). 

According to the challenge rules,
these criteria are calculated for each test image and then are averaged
on the whole test set to get the final
results. Methods are ranked by their respective JA values.

\subsection{Performance of Our Method}
We demonstrate the good performance of our proposed method by giving qualitative segmentation results and quantitative results.
\subsubsection{Qualitative Results}
\begin{figure}[htb]
\begin{minipage}[b]{.48\linewidth}
  \centering
  \centerline{\includegraphics[width=3.0cm,height=2cm]{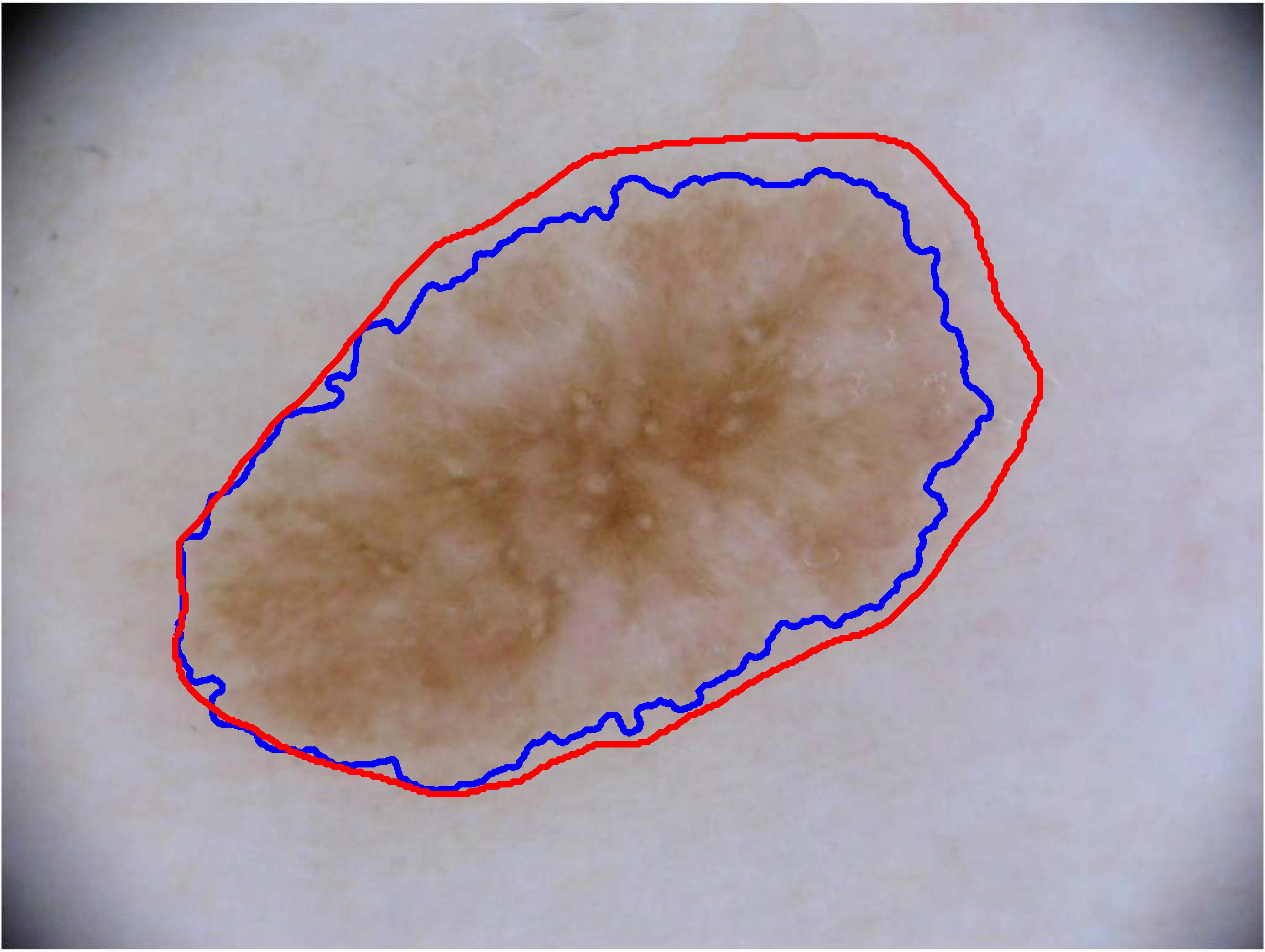}}
\end{minipage}
\begin{minipage}[b]{.48\linewidth}
  \centering
  \centerline{\includegraphics[width=3.0cm,height=2cm]{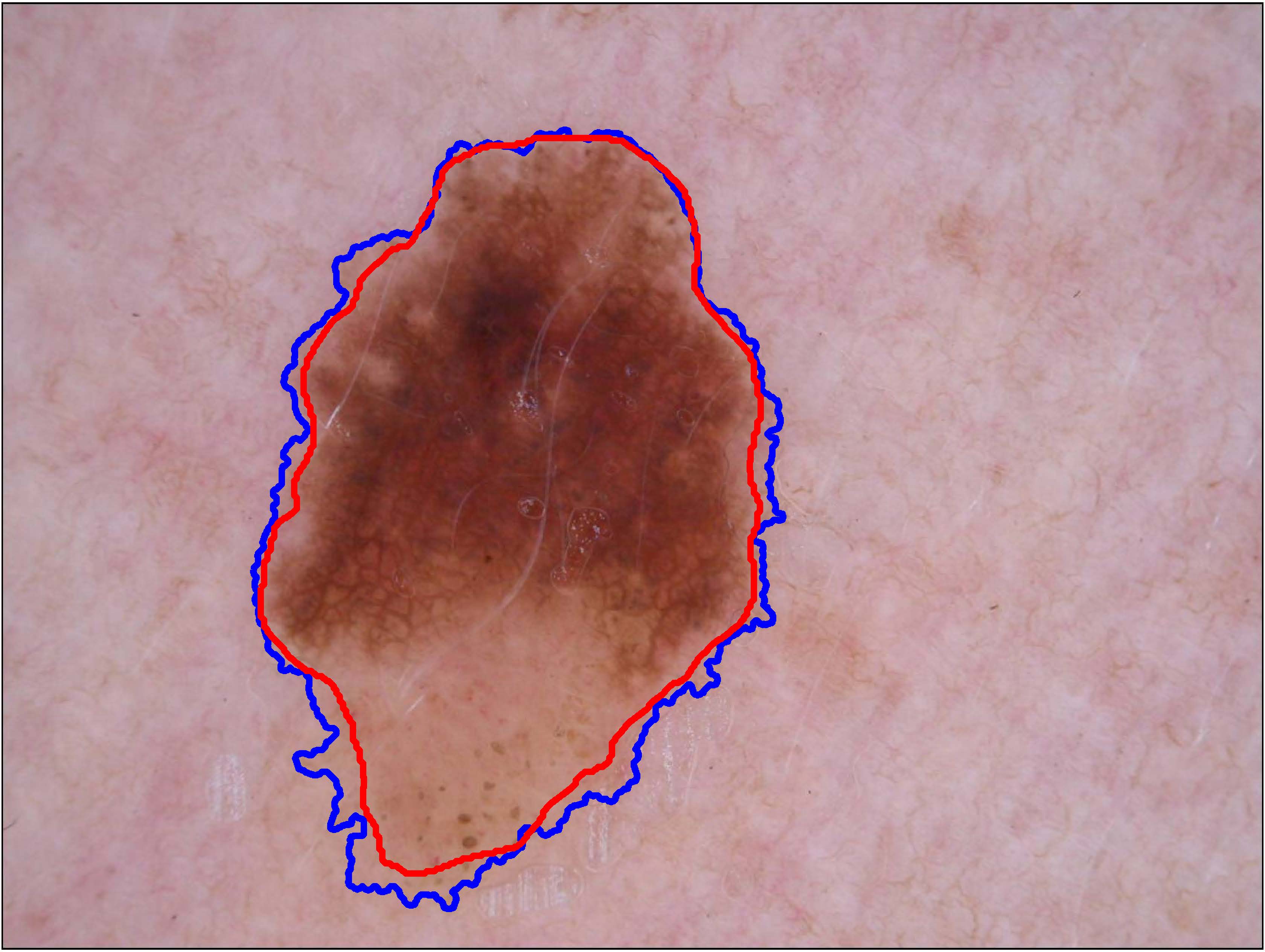}}
\end{minipage}
\hfill
\begin{minipage}[b]{0.48\linewidth}
  \centering
  \centerline{\includegraphics[width=3.0cm,height=2cm]{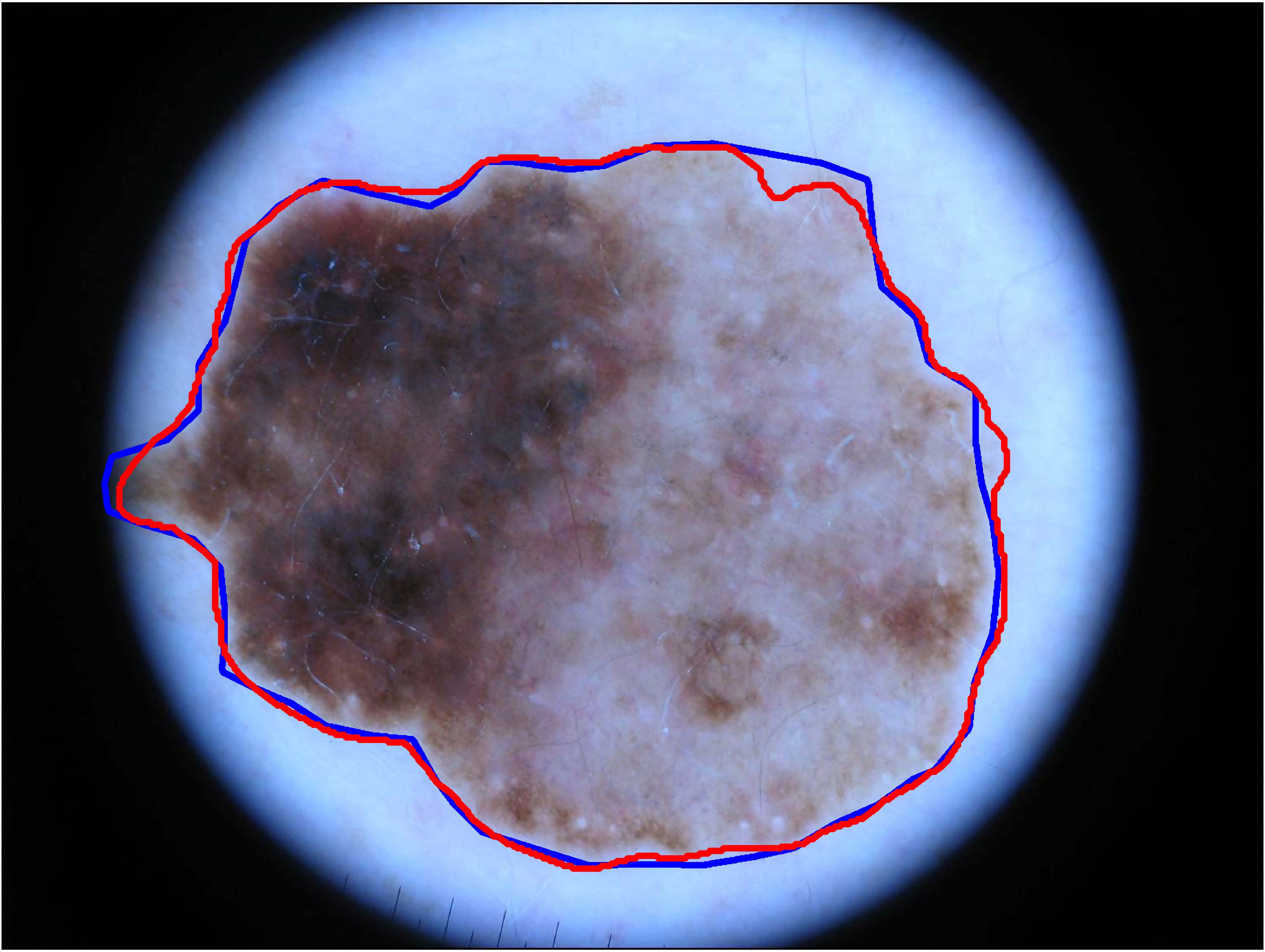}}
\end{minipage}
\begin{minipage}[b]{0.48\linewidth}
  \centering
  \centerline{\includegraphics[width=3.0cm,height=2cm]{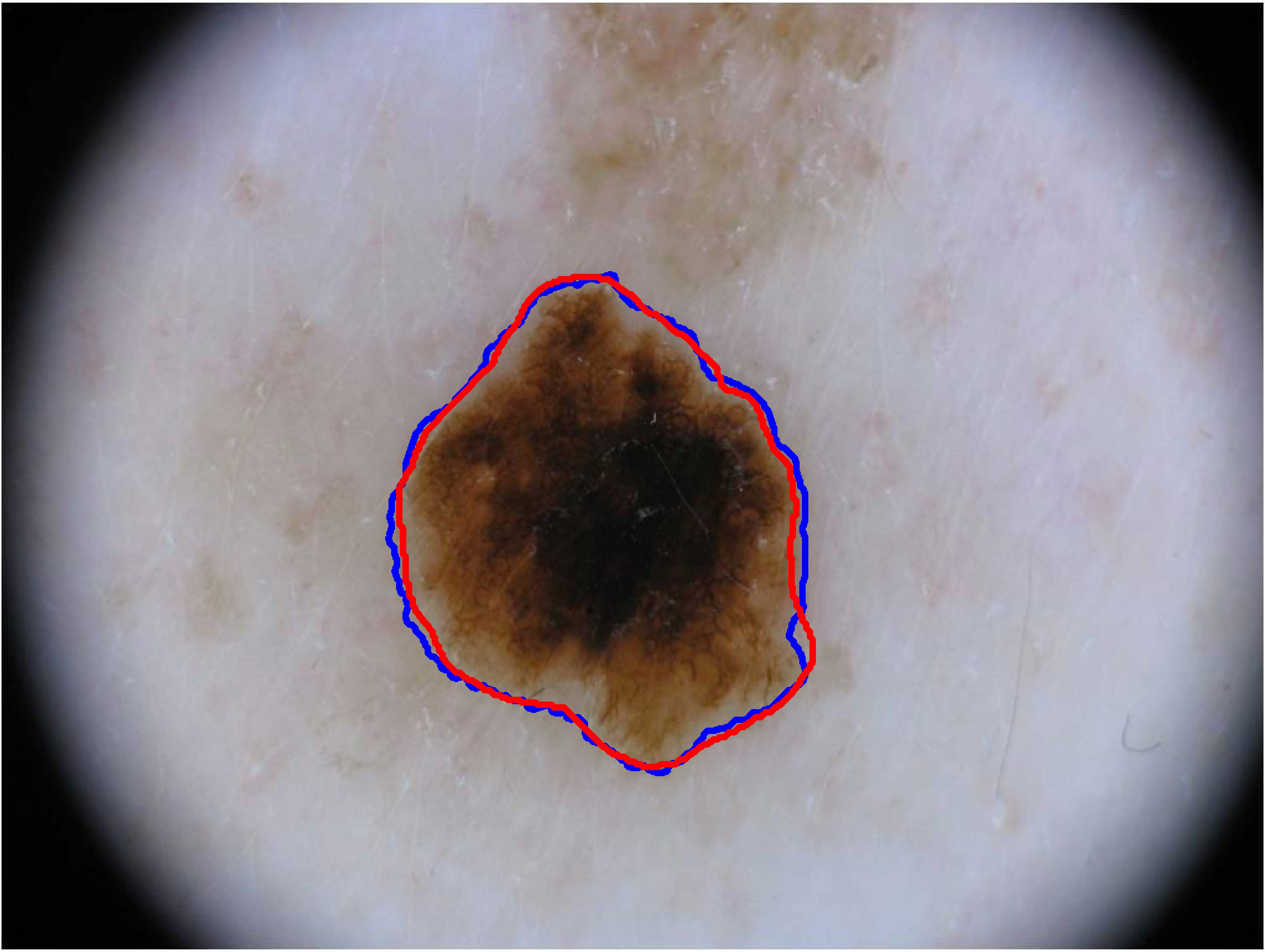}}
\end{minipage}
\begin{minipage}[b]{0.48\linewidth}
  \centering
  \centerline{\includegraphics[width=3.0cm,height=2cm]{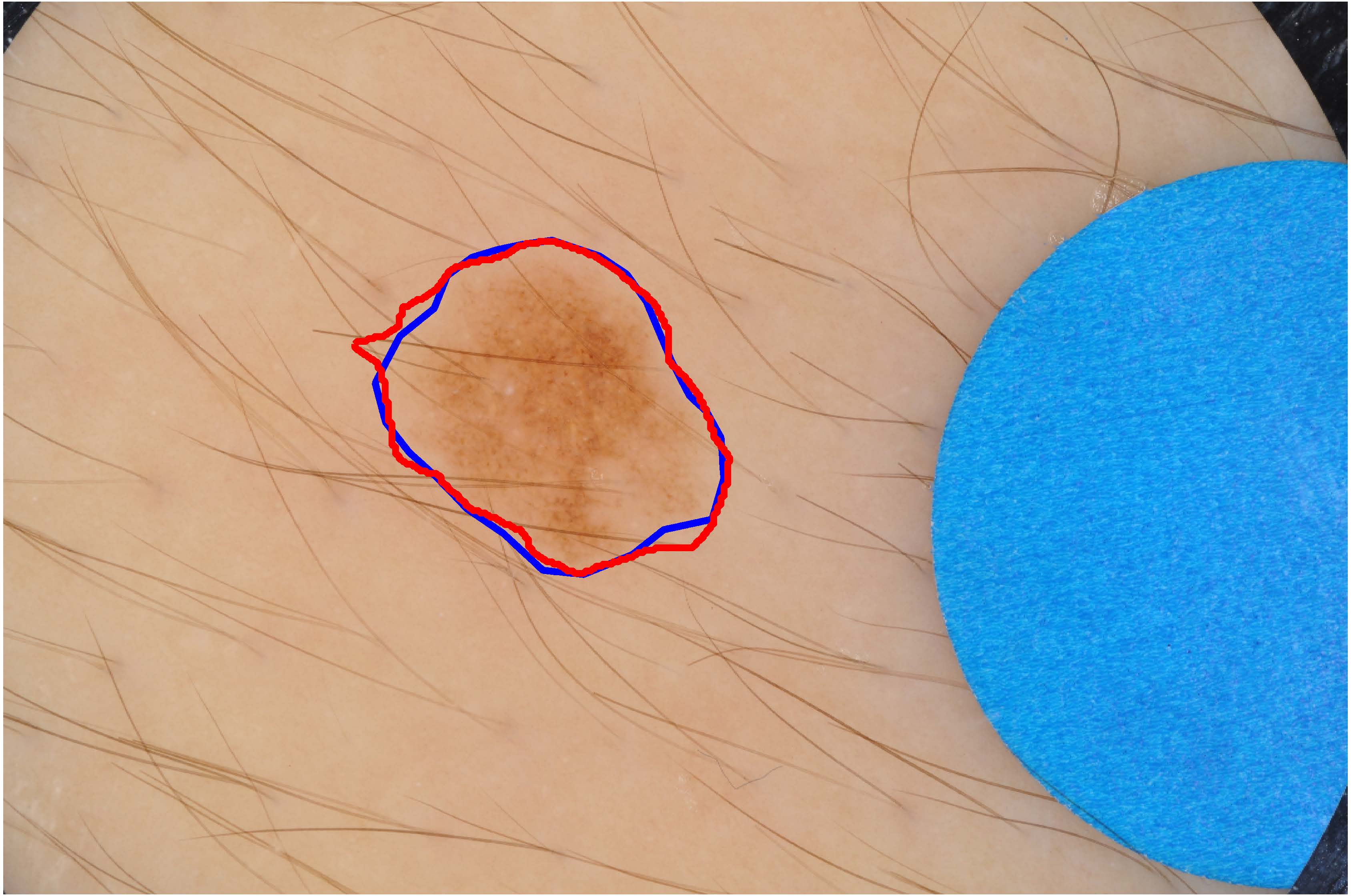}}
\end{minipage}
\hfill
\begin{minipage}[b]{0.48\linewidth}
  \centering
  \centerline{\includegraphics[width=3.0cm,height=2cm]{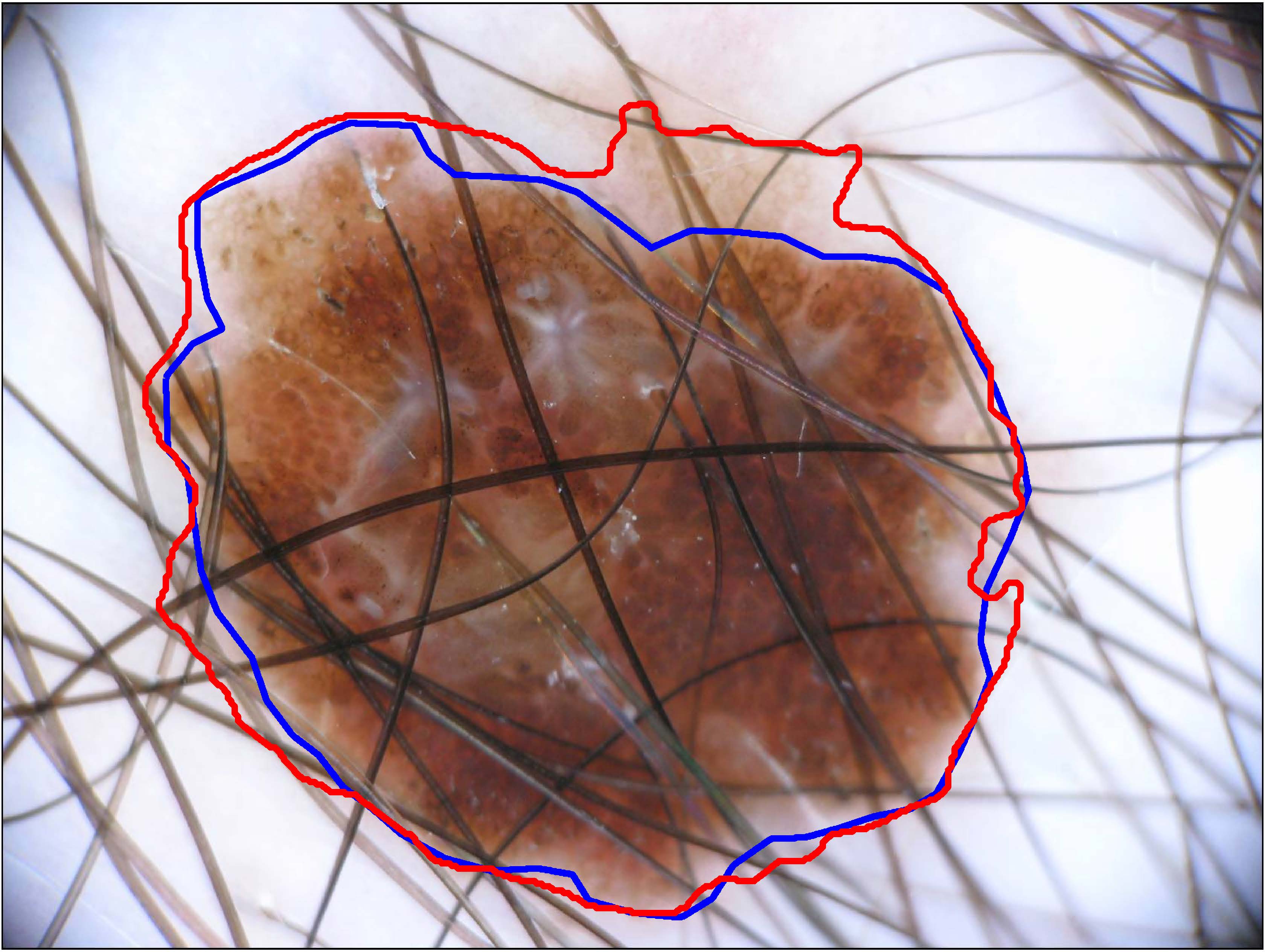}}
\end{minipage}
\caption{Example segmentation results of challenging images. Ground truth contours and our detected contours are marked in blue color and red color, respectively. First row: low contrast images; Second row: large intraclass variance; Last row: artifacts.}
\label{fig:qualitative}
\end{figure}

To show how our proposed method works on difficult images, we show our segmentation results of some challenging images in Fig. \ref{fig:qualitative} where ground truth contours and our segmented contours are marked in blue and red color, respectively. Images with low contrast, irregular shapes and artifacts are shown in the first row, second row, last row of Fig. \ref{fig:qualitative}, respectively. 

From Fig. \ref{fig:qualitative}, it can be seen that our method works very well 
in all of these challenging images.

\subsubsection{Quantitative Results}
We got and submitted preliminary results to the challenge without any pre or post processing.  

\section{DISCUSSION AND CONCLUSION}
\label{sec:discussion}
In this paper, we build a new FCNN for skin lesion segmentation.  We got and submitted preliminary results to the challenge without any pre or post processing.  The performance of our proposed method could be further improved by data augmentation and by avoiding image size reduction.
\bibliographystyle{IEEEbib}
\bibliography{strings,refs}

\end{document}